\pdfoutput=1

\documentclass[11pt]{article}

\usepackage[final]{acl}

\usepackage{times}
\usepackage{latexsym}

\usepackage[T1]{fontenc}

\usepackage[utf8]{inputenc}

\usepackage{microtype}

\usepackage{inconsolata}
\usepackage{graphicx}
\usepackage{float}
\usepackage{subfig}
\usepackage{multirow}
\usepackage{makecell}
\usepackage{booktabs}
\usepackage{amsmath}
\usepackage{amssymb}
\usepackage{stfloats}
\usepackage[normalem]{ulem}
\useunder{\uline}{\ul}{}
\title{Light-PEFT: Lightening Parameter-Efficient Fine-Tuning via Early Pruning}

\author{Naibin Gu\textsuperscript{\rm 1,2},\ Peng Fu\textsuperscript{\rm 1,2}\thanks{\ \  Corresponding author: Peng Fu.},\ Xiyu Liu\textsuperscript{\rm 1,2},\ Bowen Shen\textsuperscript{\rm 1,2},\ Zheng Lin\textsuperscript{\rm 1,2},\ Weiping Wang\textsuperscript{\rm 1} \\ 
\textsuperscript{\rm 1}Institute of Information Engineering, Chinese Academy of Sciences, Beijing, China \\
\textsuperscript{\rm 2}School of Cyber Security, University of Chinese Academy of Sciences, Beijing, China \\
  \texttt{\textrm{\{}gunaibin,fupeng,liuxiyu,shenbowen,linzheng,wangweiping\textrm{\}}@iie.ac.cn} \\
  }

\begin{document}
\maketitle
\begin{abstract}
Parameter-efficient fine-tuning (PEFT) has emerged as the predominant technique for fine-tuning in the era of large language models. However, existing PEFT methods still have inadequate training efficiency. Firstly, the utilization of large-scale foundation models during the training process is excessively redundant for certain fine-tuning tasks. Secondly, as the model size increases, the growth in trainable parameters of empirically added PEFT modules becomes non-negligible and redundant, leading to inefficiency. To achieve task-specific efficient fine-tuning, we propose the Light-PEFT framework, which includes two methods: Masked Early Pruning of the Foundation Model and Multi-Granularity Early Pruning of PEFT. The Light-PEFT framework allows for the simultaneous estimation of redundant parameters in both the foundation model and PEFT modules during the early stage of training. These parameters can then be pruned for more efficient fine-tuning. We validate our approach on GLUE, SuperGLUE, QA tasks, and various models. With Light-PEFT, parameters of the foundation model can be pruned by up to over 40\%, while still controlling trainable parameters to be only 25\% of the original PEFT method. Compared to utilizing the PEFT method directly, Light-PEFT achieves training and inference speedup, reduces memory usage, and maintains comparable performance and the plug-and-play feature of PEFT\footnote{Our code is available at \url{https://github.com/gccnlp/Light-PEFT}.}. 

\end{abstract}

\section{Introduction}
Large-scale pre-trained language models have demonstrated outstanding performance in various natural language processing domains \citep{DBLP:journals/corr/abs-1907-11692,DBLP:conf/nips/BrownMRSKDNSSAA20,Touvron2023LLaMAOA,OpenAI2023GPT4TR}. Along with the performance improvements, the scale of model parameters continues to grow, making the cost of fine-tuning models increasingly expensive. Moreover, the practice of maintaining a separate copy of the large model for each task in conventional fine-tuning incurs substantial storage costs.

To address these challenges, parameter-efficient fine-tuning (PEFT) has been proposed: freezing most parameters of the foundation model and fine-tuning only a small number of parameters \citep{DBLP:conf/icml/HoulsbyGJMLGAG19,li-liang-2021-prefix,liu-etal-2022-p,DBLP:conf/iclr/HuSWALWWC22}, thereby reducing the computational resource requirements during training and performing nearly full-parameter fine-tuning. In addition, this technique eliminates the need to save an entire model copy for each task. During inference, task-specific models can be obtained by switching directly to the appropriate parameter-efficient module for the given task.

However, the training efficiency of existing PEFT methods still needs improvement. The first problem lies in the excessive redundancy of using a large-scale foundation model during fine-tuning for specific tasks, which results in substantial computational costs. A typical strategy is to integrate PEFT with  quantization \citep{dettmers2023qlora,kim2023memoryefficient}. Nonetheless, these methods only quantize parameters to low-bit in memory, without reducing the number of parameters and they still need to be dequantized to high-bit during training, leading to wasted training time. Another more direct approach for reducing parameters is model structured pruning \citep{DBLP:journals/corr/abs-2211-10155,DBLP:journals/corr/abs-2307-07705}. However, most methods mainly focus on the inference efficiency of the model, which means they may result in higher training costs. 

The second problem is that as the size of the foundation model increases, the number of parameters in added trainable modules also increases significantly. This introduces a lot of redundancy in trainable parameters, leading to inefficiency in fine-tuning. For instance, the commonly used methods LoRA \citep{DBLP:conf/iclr/HuSWALWWC22} and QLoRA \citep{dettmers2023qlora} empirically insert the low-rank modules onto fixed weight. However, there is no need to uniformly add trainable modules of the same rank to all weights for fine-tuning each task. An improved approach is the dynamic rank method \citep{DBLP:conf/iclr/ZhangCBH0CZ23,valipour-etal-2023-dylora,ding-etal-2023-sparse}, which adaptively allocates module parameters by progressively calculating the importance of the rank during training. However, these methods require continuous estimation during training and show limited improvement in actual training efficiency.

In this paper, we introduce a novel framework named Light-PEFT, which aims to enhance the efficiency of the PEFT technique during fine-tuning. The framework consists of two methods: Masked Early Pruning of Foundation Model and Multi-Granularity Early Pruning of PEFT. In the early training stage, Light-PEFT estimates redundant parameters in both the foundation model (heads and intermediate dimensions) and the PEFT modules (module importance and rank importance) simultaneously. Structured pruning is used to eliminate this redundancy, resulting in a lighter foundation model and PEFT module for more efficient fine-tuning.

To validate the effectiveness of our Light-PEFT framework, we conduct extensive evaluations on various foundation models (RoBERTa, OPT-1.3B, OPT-6.7B), different PEFT structures (LoRA, Adapter), and on diverse benchmarks (GLUE, SuperGLUE, and question-answering tasks). The empirical results  indicate that the proposed Light-PEFT framework outperforms other baseline methods in performance. It significantly improves training efficiency that reduces training memory usage by 39\% and accelerates training to 1.6$\times$. Additionally, the Light-PEFT framework improves inference efficiency that reduces inference memory by 48\% and increases inference speed to 1.6$\times$.
\section{Related Works}
\subsection{Parameter-Efficient Fine-Tuning}
Parameter-Efficient Fine-Tuning has been proposed to reduce the computational cost of fine-tuning entire model parameters \citep{DBLP:conf/icml/HoulsbyGJMLGAG19,li-liang-2021-prefix,DBLP:conf/iclr/HuSWALWWC22}. Following works aim to further improve the efficiency of PEFT.

\noindent\textbf{Improvements to the PEFT module.} The motivation behind of this category of methods is that previous works often insert trainable modules empirically, resulting in uniform ranks for all inserted modules that are not parameter-efficient. AdaLoRA \citep{DBLP:conf/iclr/ZhangCBH0CZ23} proposes obtaining the optimal rank for each module by iteratively pruning ranks during training. DyLoRA \citep{valipour-etal-2023-dylora} achieves this through dynamic training on a range of ranks. AutoPEFT \citep{DBLP:journals/corr/abs-2301-12132} automatically selects PEFT configurations through Bayesian optimization. Recently, SoRA \citep{ding-etal-2023-sparse} introduces masks on the ranks and gradually makes each module sparse. However, all of these methods gradually calculate the rank allocation during training, which does not improve the actual training efficiency in fine-tuning. Our method estimates the rank allocation for each module in the early stage of training and utilizes the pruned parameter-efficient modules to improve training efficiency during fine-tuning.

\noindent\textbf{Improvements to the training paradigm of PEFT.} To enhance training efficiency, one idea is to further reduce the memory footprint during training. QLoRA \citep{dettmers2023qlora} and PEQA \citep{kim2023memoryefficient} reduce memory usage by quantizing the foundation model, while LST \citep{sung2022lst} and MEFT \citep{liao2023make}, respectively alleviate the memory footprint of intermediate activations in the foundation model through methods ladder side-tuning and reversible structures. Our approach is orthogonal to these methods from a memory perspective and can be combined with them. We explore early-stage pruning of the foundation model to reduce memory usage. Moreover, our approach can lower computational costs, speed up training, and improve inference efficiency.

Combining PEFT with pruning, most of works focus on improving inference efficiency. PST \citep{li_parameter-efficient_2022} and DSEE \citep{chen-etal-2023-dsee} propose combining unstructured pruning and PEFT, which hardly achieves acceleration on practical hardware. SPAs \citep{DBLP:journals/corr/abs-2211-10155} integrates structured pruning of the foundation model with PEFT, while CPET \cite{DBLP:journals/corr/abs-2307-07705} proposes distilling knowledge into PEFT modules simultaneously with pruning to reduce performance degradation. Concurrently to our works, APT \citep{DBLP:journals/corr/abs-2401-12200} reduces the training cost of the CPET method, presenting more efficient distillation and pruning. However, these methods, including APT, still require higher training time and memory costs compared to the original PEFT methods. Our approach aims to reduce the original PEFT training costs, including speed and memory, by employing early-stage structured pruning to train a non-redundant PEFT model efficiently, while improving inference efficiency simultaneously.

\subsection{Structured Pruning of Models}
Model pruning has been proposed to compress redundant parameters in models \citep{DBLP:conf/nips/CunDS89,kurtic-etal-2022-optimal,liu-etal-2022-learning-win,ma2023llmpruner}, with structured pruning being the most straightforward method to achieve acceleration on actual hardware. For the structured pruning of Transformer models, the focus lies in pruning components of the model, such as attention heads and feed-forward dimensions \citep{DBLP:conf/aaai/LiuLY21, xia-etal-2022-structured,tao-etal-2023-structured,xia2024sheared}. However, most structured pruning works require additional costs during training to obtain smaller and more accurate models for inference efficiency. In terms of training efficiency, \citet{you2020drawing} base on the lottery ticket hypothesis \citep{frankle2018the} and discover the existence of early winning tickets in DNN models, allowing early pruning to enhance subsequent training efficiency. Subsequently, \citet{chen-etal-2021-earlybert} identify early tickets in BERT models \citep{devlin-etal-2019-bert} to enhance the efficiency of BERT's pre-training and fine-tuning. We follow these works and explore early pruning in parameter-efficient fine-tuning and generative foundation models.
\section{Preliminaries}
\subsection{Parameter-Efficient Fine-Tuning}
In our framework, we choose two of the most widely used methods: Adapter \citep{DBLP:conf/icml/HoulsbyGJMLGAG19} and LoRA \citep{DBLP:conf/iclr/HuSWALWWC22} to validate our approach.

\noindent\textbf{Adapter.} For each layer in the foundation model, including the attention sub-layer and the feed-forward sub-layer, Adapter inserts a trainable MLP module after each sub-layer. It consists of a down-projection layer $W_{down}\in{R^{d\times r}}$, followed by a non-linear activation function $f$, and finally an up-projection layer $W_{up}\in{R^{r\times d}}$, where $d$ is the hidden size of the foundation model, and $r$ is the bottleneck dimension in the trainable module, with $r\ll d$. The Adapter method can be formulated as follows:
\begin{equation}
    h\leftarrow h+f(hW_{down})W_{up}
\end{equation}
where $h$ is the output of the inserted sub-layer.

\noindent\textbf{LoRA.} For each linear weight matrix $W\in{R^{d\times d}}$ in the foundation model, the LoRA method adds trainable MLP modules in parallel to $W$. The trainable module includes a down-projection layer $W_{down}$ and an up-projection layer $W_{up}$. The LoRA method can be be formulated as follows:
\begin{equation}
    h\leftarrow h+s\cdot XW_{down}W_{up}
\end{equation}
where $X$ represents the input to the linear weight matrix $W$ and $s$ is a hyper-parameter scaling factor.
\subsection{PEFT Training Efficiency}
In this section, we present observations on the training efficiency of PEFT. We utilize LoRA to observe the results on two foundation models, RoBERTa \citep{DBLP:journals/corr/abs-1907-11692} and OPT \citep{Zhang2022OPTOP}. For training samples, we set the length to 128 with a batch size of 32 and the time is the sum of 10 batches. All tests are conducted on a single NVIDIA RTX 3090 GPU.

\begin{figure}[H]
        \centering
	\subfloat[Training Time]{
	\includegraphics[width=0.48\linewidth,trim=10 10 10 10,clip]{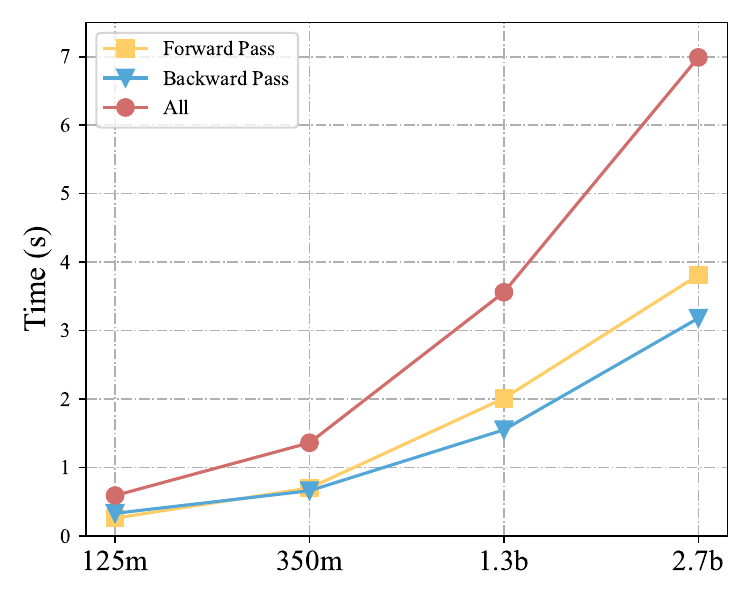}} 
        \subfloat[Memory Usage]{
	\includegraphics[width=0.48\linewidth,trim=10 10 10 10,clip]{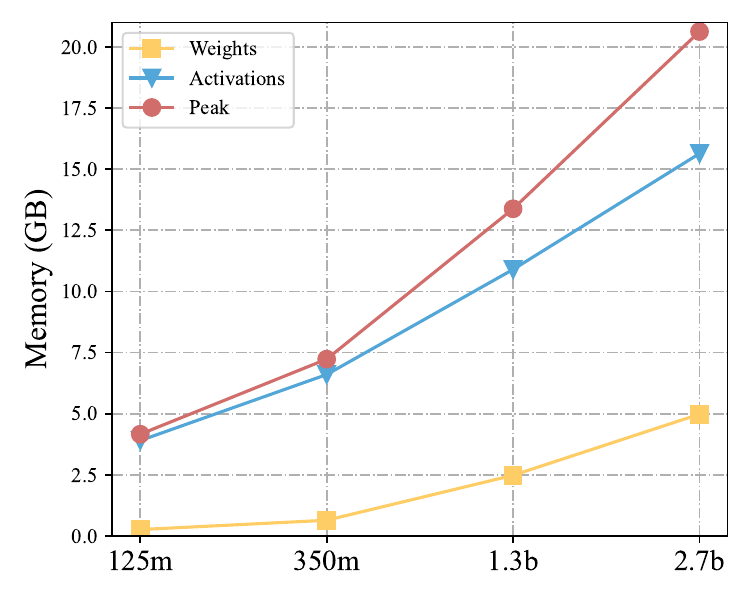}}
\caption{Impact of the foundation model size on training efficiency. The experiments are conducted on OPT models (FP16). As the size of foundation models increases, the time for forward and backward pass during training and the required memory significantly increase.}
\label{PilotA}
\end{figure}

\noindent\textbf{The impact of foundation models size.} From the perspective of training speed (Figures~\ref{PilotA}a), PEFT methods reduce the gradient computation time, so the forward pass time gradually surpasses the backward pass time. Nonetheless, the forward calculation is still unchanged and needs to use all model parameters to propagate the state forward and backpropagate the loss through the entire model, becoming slower as the model size increases. From the memory perspective (Figure~\ref{PilotA}b), although PEFT techniques reduce the memory consumption of optimizer states and gradients, the model weights and intermediate activations still occupy a significant amount of memory during training. Compressing the foundation model to a smaller size can better alleviate it.  This highlights the importance of reducing the parameter redundancy of the foundation model for training efficiency.

\begin{figure}[H]
        \centering
	\subfloat[Module rank]{
	\includegraphics[width=0.48\linewidth,trim=10 10 10 10,clip]{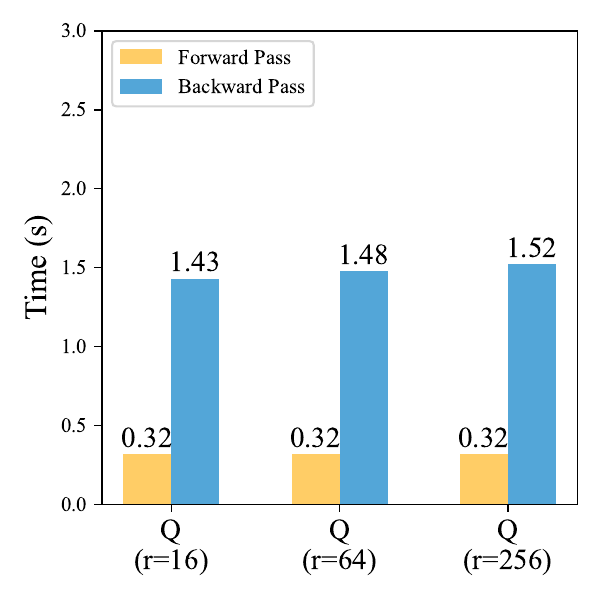}} 
        \subfloat[Number of modules]{
	\includegraphics[width=0.48\linewidth,trim=10 10 10 10,clip]{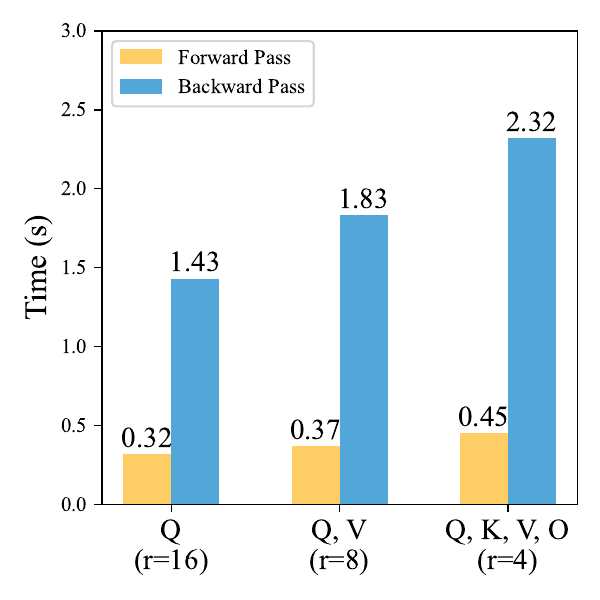}}
\caption{Impact of intra-module rank and the number of PEFT modules on training speed. The experiments are conducted on RoBERTa-Large (FP32). Q, K, V, O denote the Query, Key, Value and Output matrices in the foundation model's attention sub-layer. (a) Keeping the same number of modules and increasing the rank results in a relatively small change in pass time. (b) Increasing the number of modules while keeping the same trainable parameters leads to a significant change in pass time.}
\label{PilotB}
\end{figure}

\noindent\textbf{The impact of PEFT modules.} We explore the impact of intra-module rank and the number of PEFT modules on training efficiency. From the perspective of training speed, Figure~\ref{PilotB}a presents experiments where we keep same modules and only increase the rank. Figure~\ref{PilotB}b shows experiments where we keep the same trainable parameter, adding structured PEFT modules to different weights. It can be observed that when increasing the number of PEFT modules compared to varying the rank, both forward and backward times significantly increased. This indicates that, during training, the impact on speed of adding more structured PEFT modules is significantly larger than that of increasing in rank of a single structured module. From a memory perspective, the trainable parameters affect the memory consumption of optimizer states and gradients during training. As the size of the foundation model increases, the redundancy introduced by empirically adding trainable parameter modules impacts training efficiency.
\section{Methodology}
\subsection{Overview of Light-PEFT}
\begin{figure*}[ht]
    \centering
    \includegraphics[width=2\columnwidth]{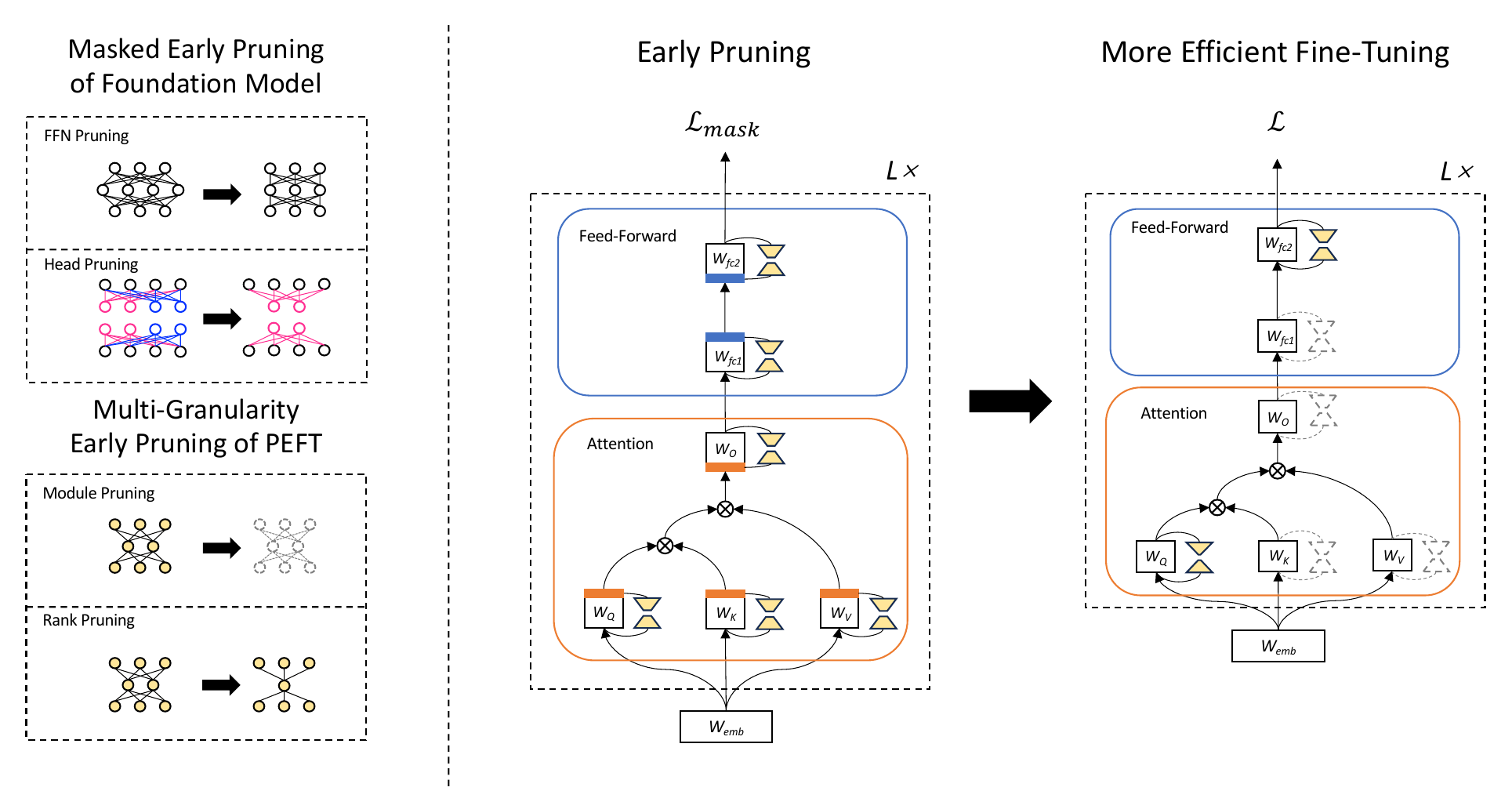}
    \caption{Illustration of Light-PEFT. The left side shows the two methods in Light-PEFT. On the right side is an illustration of the paradigm. Firstly, both methods simultaneously estimate redundancies during the early-stage of training. After estimation,  Light-PEFT prunes redundancies in both, obtaining a non-redundant foundation model and PEFT modules for more efficient fine-tuning.}
    \label{Light-PEFT}
\end{figure*}

Our goal is to eliminate parameters redundancies in the early stage, thereby reducing the computational costs of fine-tuning. Thus, we propose the Light-PEFT framework as shown in Figure \ref{Light-PEFT}, which consists of two methods: Masked Early Pruning of Foundation Model to reduce the redundancy of the foundation model and Multi-Granularity Early Pruning of PEFT to reduce the redundancy of the trainable parameters. First, both methods in our framework simultaneously estimate redundancies during the early stage of training, where the total training steps are denoted as $t$, and the estimation for early pruning steps denoted as $t'$, $t' \ll t$. After estimation,  we prune redundancies in both, thus obtain a non-redundant foundation model and PEFT modules for more efficient fine-tuning. Besides the PEFT parameters, we only need to additionally save mask vectors, which are much smaller than PEFT modules, to record the pruning index of the foundation model. During inference, our method allows the masks and PEFT modules to be easily changed, maintaining the plug-and-play feature.
\subsection{Masked Early Pruning of Foundation Model}
\label{method-1}
A typical Transformer model \citep{DBLP:conf/nips/VaswaniSPUJGKP17} consists of $L$ layers, each with a multi-head attention (MHA) sub-layer and a feed-forward network (FFN) sub-layer. A MHA sub-layer contains $N_{H}$ attention heads and weight matrices $W_{Q}^{(i)}, W_{K}^{(i)}, W_{V}^{(i)} \in \mathbb{R}^{d\times d_{H}}$, $ W_{O} \in \mathbb{R}^{d\times d}$ are used for query, key, value and output, where $d$ is the hidden size and $d_{H}=d/N_{H}$ is the hidden size of a head. In parameter-efficient fine-tuning, the weights of the foundation model are frozen, and we add the PEFT module's $\Delta W$ to these matrices. Taking the LoRA module as an example, for an input $X$ the output of the MHA is calculated as follows:
\begin{equation}
\begin{aligned}
    head^{(i)}&=(W_{Q}^{(i)}+{\Delta W}_{Q}^{(i)}, \\
    &W_{K}^{(i)}+{\Delta W}_{K}^{(i)}, W_{V}^{(i)}+{\Delta W}_{V}^{(i)},X)
\end{aligned}
\end{equation}
\begin{equation}
\begin{aligned}
    {\rm MHA}(X)&=Concat(head^{(1)},..., \\
    &head^{(N_{H})})(W_{O}+{\Delta W}_{O})
\end{aligned}
\end{equation}
To identify redundancy in attention heads, we introduce a trainable scalar mask $m_A$ in each layer's MHA sub-layer. Now the MHA becomes:
\begin{equation}
\begin{aligned}
    head^{(i)}&=m_{A}^{(i)}\cdot(W_{Q}^{(i)}+{\Delta W}_{Q}^{(i)}, \\
    &W_{K}^{(i)}+{\Delta W}_{K}^{(i)}, W_{V}^{(i)}+{\Delta W}_{V}^{(i)},X)
\end{aligned}
\end{equation}

For a FFN sub-layer, which contains activation function ${\rm Act}(\cdot)$ and weight matrices $W_{fc1}$ and $W_{fc2}$ which denote up-projection and down-projection. With PEFT modules, for an input $X$ the output of the FFN is calculated as follows:
\begin{equation}
\begin{aligned}
    {\rm FFN}(X)=&{\rm Act}(X(W_{fc1}+{\Delta W_{fc1}})) \\
    &\cdot(W_{fc2}+{\Delta W_{fc2}})
\end{aligned}
\end{equation}
We also introduce a trainable scalar mask $m_F$ in each layer's FFN sub-layer to eliminate redundancy in intermediate dimension. Now the FFN become:
\begin{equation}
\begin{aligned}
    {\rm FFN}(X)=&{\rm Act}(X(W_{fc1}+{\Delta W_{fc1})})\cdot{m_{F}} \\
    &\cdot(W_{fc2}+{\Delta W_{fc2}})
\end{aligned}
\end{equation}

Inspired by \citet{DBLP:conf/iccv/LiuLSHYZ17}, we then use L1 regularization to learn masks $m_A$ and $m_F$. During the mask learning, the PEFT module and the mask are trained jointly using gradient descent, which allows the mask to better present the impact of PEFT to the foundation model training on the target task. The loss function is as follows:
\begin{equation}
\mathcal{L}_{mask}=\mathcal{L}+\lambda_{A}\Vert  m_A\Vert_{1}+\lambda_{F}\Vert  m_F\Vert_{1}
\end{equation}
where $\mathcal{L}$ is the original loss in fine-tuning, $\lambda_{A}$ and $\lambda_{F}$ are hyper-parameters to control the penalty of regularization (see Appendix \ref{mask-learning-penalty} for details). The masks are initialized to 1 at the beginning of training.

After estimating, we perform structured pruning on attention heads with pruning ratio $\rho_{A}$ layer-wise and intermediate dimensions with $\rho_{F}$ globally based on the magnitudes of $m_A$ and $m_F$. 
\subsection{Multi-Granularity Early Pruning of PEFT}
\label{method-2}
In comparison to the fine-grained sparsity (i.e. rank allocation) that is the focus of most previous works \citep{DBLP:conf/iclr/ZhangCBH0CZ23,valipour-etal-2023-dylora}, our preliminary observation also confirms the significance of coarse-grained module pruning for training speed. Therefore, we propose multi-granularity PEFT pruning to consider both aspects simultaneously. Furthermore, we perform pruning PEFT in the early stage to maximize efficiency during training.
\subsubsection{Modules Pruning}
To achieve coarse-grained module pruning, we begin with the original design of PEFT, where we believe that the importance of a module is primarily determined by the change it brings to the original information. Specifically, for the LoRA method, we add a trainable module $ W_{down}W_{up}$ on the weight $W$. Thus, given an input $X$, the importance ratio $I_M$ is defined as:
\begin{equation}
I_M=\frac{\Vert  {X\cdot W_{down}W_{up}}\Vert_{2}}{\Vert  {X\cdot W}\Vert_{2}}
\end{equation}
where ${\Vert  {\cdot}\Vert_{2}}$ represents the L2 norm, measuring the magnitude of the vector output from the PEFT module. Because one of the weight matrices in the PEFT module, such as $W_{up}$ in the LoRA method, is typically initialized to zero. Therefore, during training, the ratio of the output magnitude of the LoRA module to the weight $W$'s output magnitude indicates the importance of the changes required by the module added at that position.

For the Adapter method, a trainable module is added after a sub-layer. Given the output $h$ of the previous sub-layer, the importance ratio $I_M$ is defined as:
\begin{equation}
I_M=\frac{\Vert  {f(hW_{down})W_{up}}\Vert_{2}}{\Vert  {h}\Vert_{2}}
\end{equation}
where $I_M$ represents the change in information of the Adapter module on the output information $h$ of the previous sub-layer.

In the implementation, to better estimate the importance of all added positions for the LoRA method, we add LoRA modules on all weights of the foundation model. This may result in higher costs compared to the original LoRA in the short term, but our early estimation steps are significantly smaller than the total training steps, allowing for a substantial reduction in total costs. For the Adapter method, we follow the original approach by adding them after both the MHA and FFN sub-layers. After estimation, we use $I_M$ to globally prune the entire PEFT modules with the pruning rate $\rho_{M}$.
\subsubsection{Ranks Pruning}
In addition to coarse-grained pruning, we further perform fine-grained pruning on the rank of the modules. This allows us to reduce more trainable parameters and enhance training efficiency. Our motivation is based on the fact that not all modules require the same rank allocation. 
To eliminate redundant ranks, we use the first-order Taylor expansion \citep{molchanov2017pruning} to estimate the importance $I_{W_{i,j}}$ of each parameter connected to the rank in the PEFT module:
\begin{equation}
I_{W_{i,j}}=\left| {\frac{\partial{\mathcal{L}}}{\partial{W_{i,j}}}{W_{i,j}}} \right|
\end{equation}
where $W_{i,j}$ represents the i-th row and j-th column of parameters in $W_{down}$ or $W_{up}$ of the PEFT module. The importance of the rank $I_R$ is the sum of the importance $I_{W_{i,j}}$ of all parameters corresponding to the rank in the column of $W_{down}$ and the row of $W_{up}$. After estimation, we globally prune the unimportant ranks with pruning rate ${\rho}_{R}$.

\section{Experiments}
\subsection{Experimental Setup}
\noindent\textbf{Datasets and evaluation.} We conduct experiments on eight natural language understanding (NLU) tasks from GLUE \cite{DBLP:conf/iclr/WangSMHLB19} and SuperGLUE \cite{DBLP:conf/nips/WangPNSMHLB19} and six question-answering (QA) tasks. Because our goal is to enhance training efficiency, training on small datasets does not hold much significance. As a result, we choose four larger datasets from GLUE including MNLI \cite{williams-etal-2018-broad}, QNLI \cite{rajpurkar-etal-2016-squad}, QQP\footnote{https://data.quora.com/First-Quora-Dataset-Release-Question-Pairs}, and SST-2 \cite{socher-etal-2013-recursive}, and four larger datasets from SuperGLUE comprising ReCord \cite{DBLP:journals/corr/abs-1810-12885}, WiC \cite{pilehvar-camacho-collados-2019-wic}, BoolQ \cite{clark-etal-2019-boolq}, and MultiRC \cite{khashabi-etal-2018-looking}. For MNLI, we report accuracy on the matched validation set. For QNLI, QQP, SST-2, WiC and BoolQ we report accuracy. For ReCord we report F1 and for MultiRC we report F1 over all answer-options (F1$_{a}$). The QA tasks include OpenBookQA \cite{mihaylov-etal-2018-suit}, PIQA \cite{DBLP:conf/aaai/BiskZLGC20}, ARC-Easy and ARC-Challenge \cite{DBLP:journals/corr/abs-1803-05457}, SciQ \cite{welbl-etal-2017-crowdsourcing} and WebQuestions \cite{berant-etal-2013-semantic}. We report accuracy on all QA tasks by lm-evaluation-harness \cite{eval-harness}.

\noindent\textbf{Baselines.} We use RoBERTa-Large for NLU tasks, OPT-1.3B and OPT-6.7B for QA tasks as foundation models. We choose several baselines to verify the effectiveness of our
method. \textbf{Full-FT} is the conventional approach for fine-tuning. \textbf{Adapter} \cite{DBLP:conf/icml/HoulsbyGJMLGAG19} and \textbf{LoRA} \cite{DBLP:conf/iclr/HuSWALWWC22} are original structures we used in our framework. \textbf{LayerDrop} \cite{DBLP:conf/iclr/FanGJ20} is a strong baseline method that enhances training efficiency by dynamically dropout layers during training. We re-implement it combining with LoRA method. \textbf{LST} \cite{sung2022lst} improves model training efficiency by avoiding backpropagation in the foundation model. \textbf{Offsite-Tuning} \cite{DBLP:journals/corr/abs-2302-04870} uses a emulator derived from the foundation model for efficient training, and replaces the emulator's layers back into the foundation model for inference. \textbf{LLM-Pruner} \cite{ma2023llmpruner} prunes model on small amount of task-agnostic corpora and restores performance using LoRA, thereby improving training efficiency. We re-implement their original task-agnostic pruning and add a task-specific pruning implementation using 1k random samples from task data.

\noindent\textbf{Implementation.} For the GLUE benchmark, we control the estimation steps for early pruning to be around 5\% of the total training steps. For the more challenging SuperGLUE benchmark, we set the estimation steps to be within 10\%. For QA tasks, we uniformly use 10\% of the training steps. Please refer to the Appendix \ref{exp-setup} for detailed pruning settings, as well as other training details. 
\subsection{Experimental Results}
\label{main-results}
\begin{table*}[ht]
\centering
\scalebox{0.9}{
\begin{tabular}{@{}ccccccccc@{}}
\toprule
\multirow{2}{*}{Method} &
  \multirow{2}{*}{\makecell[c]{\#Trainable \\Params}} &
  \multirow{2}{*}{\makecell[c]{\#Foundation \\Model Params}} &
  \multicolumn{5}{c}{GLUE} &
  \multirow{2}{*}{\makecell[c]{Training\\Speed up}} \\ \cmidrule(lr){4-8}
                 &        &       & MNLI & QNLI & QQP           & SST-2 & Avg. &      \\ \midrule
Full-FT               & 355.0M & 100\% & 90.4 & 94.7 & 92.2          & 96.4  & 93.4 & 0.7$\times$ \\
Adapter & 0.8M   & 100\% & 90.8 & 94.7 & 91.5          & 96.3  & 93.3 & 1$\times$   \\
LoRA    & 0.8M   & 100\% & 90.6 & 94.9 & 91.6          & 96.2  & 93.3 & 1$\times$   \\ \midrule
LayerDrop       & 0.5M   & 67\%  & 87.4 & 91.7 & 88.3          & 94.7  & 90.5 & 1.4$\times$ \\
LST              & 8.6M   & 100\% & 86.7 & 90.2 & 89.7          & 95.1  & 90.4 & 1.4$\times$ \\
Ours (Adapter)   & 0.3M   & 72\%  & 88.3 & 93.2 & \textbf{89.8} & 95.6  & 91.7 & 1.4$\times$ \\
Ours (LoRA)      & 0.3M   & 72\%  & \textbf{89.4} & \textbf{93.6} & 89.7 & \textbf{95.9} & \textbf{92.2} & 1.4$\times$ \\ \midrule
Ours (Adapter) &
  0.3M &
  67\% &
  87.6 &
  93.1 &
  89.1 &
  95.4 &
  91.3 &
  \textbf{1.6$\times$} \\
Ours (LoRA) &
  0.3M &
  67\% &
  \textbf{89.0} &
  \textbf{93.5} &
  \textbf{89.2} &
  \textbf{95.8} &
  \textbf{91.9} &
  \textbf{1.6$\times$} \\ \bottomrule
\end{tabular}
}
\caption{Results of GLUE benchmark. The training speed is measured on a single NVIDIA TITAN RTX 24GB GPU with batch size=32 and sequence length=128. Note that the speed computed here also includes the time required for estimation before pruning.}
\label{GLUE-Results}
\end{table*}
\subsubsection{Experiments on NLU Tasks}
We first evaluate our method on the GLUE benchmark. As shown in Table \ref{GLUE-Results}, we achieve comparable performance with the original method while using 72\% of the foundation model parameters (pruning 5/16 of the heads and 1/3 of the FFN intermediate dimensions) and 0.3M trainable parameters by pruning PEFT modules and ranks. This results in a 1.4$\times$ training speedup and improvements in memory usage due to pruning. Furthermore, our method outperforms the baseline methods with the same speed, having fewer trainable parameters. When increasing the pruning rate and retaining 67\% of the parameters in the foundation model, Light-PEFT achieves a 1.6$\times$ training speedup while still ensuring slightly better performance than the baselines. On the more challenging SuperGLUE benchmark, as shown in Table \ref{SuperGLUE-Results}, we prune 4/16 of the heads and 30\% of the FFN intermediate dimensions, retaining 76\% of the parameters in the foundation model and 0.3M trainable parameters. This achieves performance comparable to the original PEFT method , demonstrating the effectiveness of our method Masked Early Pruning of Foundation Model.
\begin{table}[t]
\centering
\scalebox{0.7}{
\begin{tabular}{@{}cccccccc@{}}
\toprule
\multirow{2}{*}{Method} & \multirow{2}{*}{\#T.P.} & \multirow{2}{*}{\makecell[c]{\#F.P.}} & \multicolumn{5}{c}{SuperGLUE} \\ \cmidrule(l){4-8} 
              &      &       & ReCord & WiC  & BoolQ & MultiRC & Avg. \\ \midrule
Adapter       & 0.8M & 100\% & 89.5   & 71.0   & 84.3  & 82.4    & 81.8 \\
Ours          & 0.3M & 76\%  & 86.0   & 70.1 & 81.2  & 76.0      & 78.3 \\
\midrule
LoRA          & 0.8M & 100\% & 88.3   & 72.7 & 84.1  & 82.7    & 82.0 \\
Ours          & 0.3M & 76\%  & 86.6   & 70.2 & 83.3  & 78.0      & 79.5 \\ \bottomrule
\end{tabular}
}
\caption{Results of SuperGLUE Benchmark. \#T.P. denotes the trainable parameters. \#F.P. denotes the proportion of parameters retained after pruning the foundation model.}
\label{SuperGLUE-Results}

\end{table}
\subsubsection{Experiments on QA Tasks}
For the QA tasks (Table \ref{QA-Results}), we first conduct experiments on OPT-1.3B. We prune parameters (12/32 heads and 2/5 intermediate dimensions), retaining 64\% of the foundation model parameters and 1.5M trainable parameters, achieving comparable performance to the original method. When the trainable parameter in the original LoRA method is set to 1.57M (r=8), our method outperforms the original LoRA under fewer foundation model parameters, which demonstrates the effectiveness of our method Multi-Granularity Early Pruning of PEFT.

Compared to Offsite-Tuning, our method achieves better performance without the high training costs of the distillation. Compared to LLM-Pruner, our method outperforms both task-agnostic and specific implementations, and our pruning process does not require the large model's gradients, leading to significantly reduced computational costs. Even when pruning it to 54\%, we maintain better performance than the baselines.

On the larger OPT-6.7B model, pruning more foundation model parameters than OPT-1.3B and using 5.2M trainable parameters, we achieve performance comparable to the original method. When reducing trainable parameters to 2M, our method still demonstrates good performance. These experimental results demonstrate that in QA tasks, we can use the Light-PEFT framework to remove more redundant parameters from the foundation model and trainable modules, improving training efficiency while ensuring performance.

\begin{table*}[ht]
\centering
\scalebox{0.8}{
\begin{tabular}{@{}cccccccccc@{}}
\toprule
\multirow{2}{*}{Method} & \multirow{2}{*}{\makecell[c]{\#Trainable \\Params}} & \multirow{2}{*}{\makecell[c]{\#Foundation \\Model Params}} & \multicolumn{7}{c}{QA Tasks}                 \\ \cmidrule(l){4-10} 
                     &       &       & OpenBookQA & PIQA & ARC-E & ARC-C & SciQ & WebQs & Avg.  \\ \midrule \midrule
\multicolumn{10}{c}{OPT-1.3B}                                                         \\ 
Full-FT              & 1.3B  & 100\% & 31.4       & 75.2 & 61.3  & 27.7  & 92.5 & 31.2  & 53.2 \\
Offsite-Tuning          & -                  & 100\%                                      & 29.0 & 74.5 & 59.4 & 27.8 & 92.9 & 26.2 & 51.6 \\
LoRA (r=64) & 12.6M & 100\% & 33.6       & 74.7 & 59.5  & 29.5  & 92.0 & 29.8  & 53.2 \\
LoRA (r=8)  & 1.6M  & 100\% & 29.6       & 74.6 & 59.9  & 29.1  & 93.0 & 28.7  & 52.5 \\
LLM-Pruner (ag.) & 10.6M & 70\%  & 29.0       & 72.4 & 54.0  & 24.7  & 89.2 & 20.7  & 48.3 \\
LLM-Pruner (sp.)  & 10.6M & 70\%  & 30.4       & 72.9 & 55.9  & 27.6  & 88.7 & 26.5  & 50.3 \\
Ours (LoRA)          & 1.5M  & 64\%  & 33.2       & 74.1 & 59.0  & 28.4  & 92.7 & 28.6  & 52.7 \\
Ours (LoRA)          & 1.9M  & 54\%  & 33.2       & 72.6 & 57.6  & 27.5  & 91.8 & 28.2  & 51.8 \\ \midrule \midrule
\multicolumn{10}{c}{OPT-6.7B}                                                         \\ 
Offsite-Tuning       & -     & 100\% & 33.8       & 77.7 & 66.8  & 33.9  & 91.9 & 23.9  & 54.7 \\
LoRA (r=64)           & 33.6M & 100\% & 39.2       & 78.5 & 67.5  & 36.7  & 94.0 & 38.5  & 59.1 \\
Ours (LoRA)          & 5.2M  & 52\%  & 39.4       & 74.9 & 63.4  & 32.7  & 92.9 & 35.8  & 56.5 \\
Ours (LoRA)          & 2.0M    & 52\%  & 37.2       & 76.0 & 64.4  & 31.7  & 93.3 & 34.7  & 56.2 \\ \bottomrule
\end{tabular}
}
\caption{Results of QA Tasks. Full-FT and Offsite-Tuning results are from \citet{DBLP:journals/corr/abs-2302-04870}. For the original LoRA method, we add modules (rank=64) to the Query and Value matrices to achieve results similar to Full-FT. For the LLM-Pruner method, We re-implement their original task-agnostic pruning (ag.) and add a task-specific pruning (sp.) implementation using 1k random samples from task data.}
\label{QA-Results}
\end{table*}
\begin{table}[t]
\centering
\scalebox{0.9}{
\begin{tabular}{@{}ccccc@{}}
\toprule
\multicolumn{1}{l}{\multirow{2}{*}{\makecell[c]{PEFT Pruning\\Strategy}}} & \multicolumn{2}{c}{LoRA} & \multicolumn{2}{c}{Adapter} \\ \cmidrule(l){2-5} 
\multicolumn{1}{l}{} & QNLI & SST-2 & QNLI & SST-2 \\ \midrule
all                  & 93.5 & 95.8  & 93.1 & 95.4  \\
w/o module p.        & 93.8 & 96.1  & 92.9 & 95.5  \\
w/o rank p.          & 93.8 & 95.8  & 93.2 & 95.2  \\
w/o all              & 93.6 & 95.6  & 93.0 & 95.1  \\ \bottomrule
\end{tabular}
}
\caption{Ablation Study of Multi-Granularity Early Pruning of PEFT. We investigate the results of not using coarse-grained module pruning (w/o module p.), not using fine-grained rank pruning (w/o rank p.), and not using any PEFT pruning (w/o all).}
\label{ablation}
\end{table}
\subsection{Analysis}
\subsubsection{Ablation Study}
In the Section \ref{main-results}, we have demonstrated the performance of foundation model pruning (more experiments in Appendix \ref{sparsity-impact}). Here, we conduct ablation study to examine two PEFT pruning strategies, module pruning and rank pruning (Table \ref{ablation}). Compared to not using any PEFT pruning, using module pruning or rank pruning generally improves generalization and thus enhances performance in most cases, indicating the effectiveness of the two proposed pruning strategies. 
Moreover, by combining the two pruning strategies, the model maintains a comparable level of performance despite having more pruned trainable parameters.
\begin{figure}[t]
        \centering
	\subfloat[Training Time]{
	\includegraphics[width=0.45\columnwidth,trim=10 10 10 10,clip]{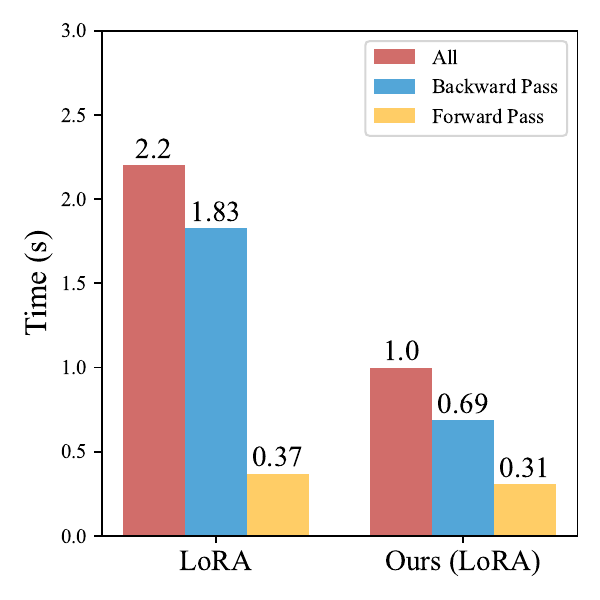}} 
        \subfloat[Memory Usage]{
	\includegraphics[width=0.45\columnwidth,trim=10 10 10 10,clip]{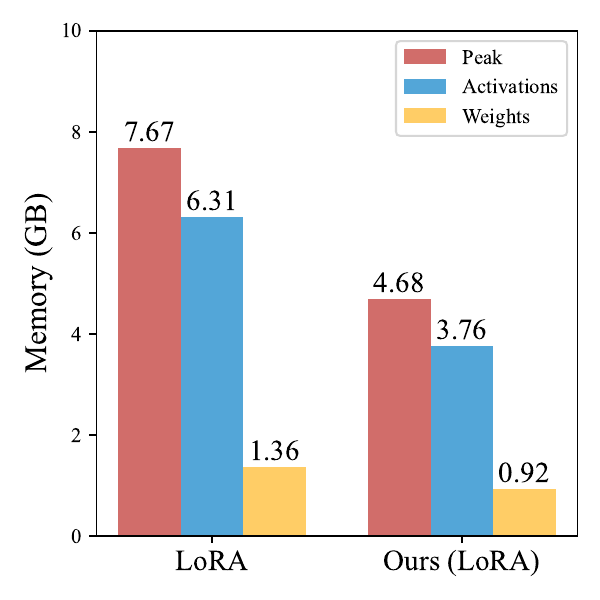}}
\caption{Training Efficiency. The experiments are conducted on RoBERTa-Large, and we set batch size=32 and sequence length=128. Our method retains 67\% of foundation model parameters and 0.3M trainable parameters.}
\label{training-eff}
\end{figure}
\subsubsection{Training and Inference Efficiency}
We validate the training and inference efficiency of our method on NVIDIA RTX 3090. In terms of training efficiency (Figure \ref{training-eff}), we conduct experiments on RoBERTa-Large, retaining 67\% of foundation model parameters and 0.3M trainable parameters that resulted in 32\% reduction in model weight memory, 40\% reduction in activations memory, and 39\% reduction in peak memory. Calculating the total time for 10 batches, we achieve 2.2$\times$ speedup in forward and backward pass time compared to the original LoRA. 

In terms of inference efficiency (Table \ref{infer-eff}), we conduct experiments on OPT-6.7B, representing widely used generative LLMs. Compared to the common practice of adding LoRA modules onto all matrices in the fine-tuning of LLMs (Vanilla), our proposed foundation model pruning and PEFT module pruning can effectively increase inference speed by up to 1.6$\times$. Additionally, foundation model pruning can effectively reduce the model loading memory usage by up to 48\%.

\begin{table}[]
\centering
\scalebox{0.73}{
\begin{tabular}{@{}ccccc@{}}
\toprule
\multirow{2}{*}{Method} & \multirow{2}{*}{\makecell[c]{\#Foundation \\Model Params}} & \multirow{2}{*}{\makecell[c]{$N_{M}$ \\($\rho_{M}$)}} & \multirow{2}{*}{\makecell[c]{Inference\\Speed Up}} & \multirow{2}{*}{\makecell[c]{Load\\Memory}} \\
&     &                     &      &       \\ \midrule
Vanilla &    100\%     & 192 (-0\%) & 1$\times$   & 12.5G \\ \midrule
\multirow{4}{*}{Light-PEFT}  &  76\% & 192 (-0\%) & 1.1$\times$ & 9.5G  \\
&    52\% & 192 (-0\%) & 1.2$\times$ & 6.5G  \\
&    52\% & 96 (-50\%) & 1.4$\times$ & 6.5G  \\
&   \textbf{52\%} & \textbf{48 (-75\%)} & \textbf{1.6$\times$} & \textbf{6.4G}  \\ \bottomrule
\end{tabular}
}
\caption{Inference Efficiency. The experiments are conducted on OPT-6.7B. $\rho_{M}$ denotes the PEFT module pruning rate, where 0\% indicates inserting LoRA modules (r=8)  onto all matrices of the foundation model. And $N_{M}$ denotes the remaining number of LoRA modules after PEFT module pruning. We set batch size=96 and max length=100.}
\label{infer-eff}
\end{table}
\section{Conclusion}
This paper introduces Light-PEFT, a novel framework designed to improve the efficiency of the PEFT technique during fine-tuning. The framework comprises two methods: Masked Early Pruning of Foundation Model and Multi-Granularity Early Pruning of PEFT. The Light-PEFT framework estimates redundant parameters in both the foundation model and PEFT modules during the early stage of training and prunes them to achieve more efficient fine-tuning. We validate our approach on GLUE, SuperGLUE, and QA tasks using various models. The experiments demonstrate that Light-PEFT achieves training and inference speedup, reduces memory usage, and maintains comparable performance.
\section*{Limitations}
Although Light-PEFT has achieved improved training and inference efficiency along with good performance, our work primarily focuses on the single-task fine-tuning scenario. A future direction worth exploring is the estimation and early pruning of redundant parameters on the multi-task learning scenario, enabling efficient fine-tuning across multiple tasks.
\section*{Ethics Statement}
The goal of our Light-PEFT framework is to enhance training efficiency and reduce computational resource costs, which has positive impacts.
\section*{Acknowledgements}
The authors thank Yuanxin Liu from Peking University and Jiaxuan Zhao from Institute of Information Engineering for the help and the anonymous reviewers for their valuable feedback on our paper.

\bibliography{anthology,custom}
\newpage
\appendix

\section{Appendix}
\label{sec:appendix}
\subsection{Details of Experimental Setup}
\label{exp-setup}
\noindent\textbf{Hardware.} We use NVIDIA TITAN RTX and NVIDIA RTX 3090 for NLU experiments and experiments using OPT-1.3B in QA Tasks. Additionally, we use NVIDIA A800 for experiments using OPT-6.7B in QA Tasks.

\noindent\textbf{Implementation.} The implementation of Light-PEFT is based on Transformers \citep{wolf-etal-2020-transformers}, LLM-Adapters \citep{hu-etal-2023-llm}, and EarlyBERT \citep{chen-etal-2021-earlybert}. The data processing for SuperGLUE and QA tasks follows \citet{liu-etal-2022-p} and \citet{DBLP:journals/corr/abs-2302-04870}, respectively.

\noindent\textbf{Hyper-parameters.} We use AdamW as the optimizer for training. Other detailed settings for NLU tasks are provided in Table \ref{glue-setting}, while the settings for QA tasks can be found in Table \ref{qa-setting-13} and Table \ref{qa-setting-67}.
\subsection{The impact of the pruning rate on the foundation model.}
\label{sparsity-impact}
\begin{figure}[H]
    \centering
    \includegraphics[width=0.8\columnwidth]{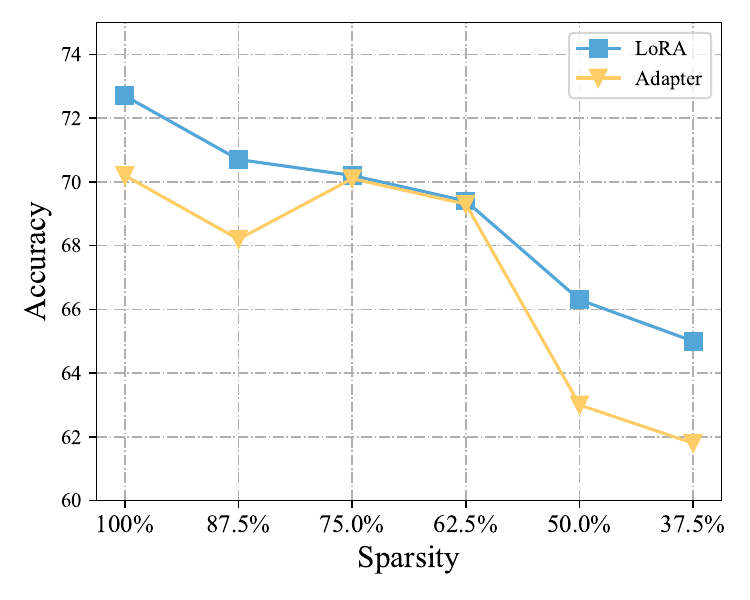}
    \caption{The impact of the pruning rate on the foundation model.}
    \label{Appendix-sparsity}
\end{figure}

We analyze the impact of different foundation model pruning rates on performance on the WiC dataset (Figure \ref{Appendix-sparsity}). It is observed that within a certain range (above 62.5\%), pruning results in a relatively minor decrease in performance. However, once this threshold is exceeded, a significant performance decline occurs, demonstrating that pruning within this range removes redundant parameters.
\subsection{The impact of the estimation steps of early pruning}
\begin{figure}[H]
    \centering
    \includegraphics[width=0.8\columnwidth]{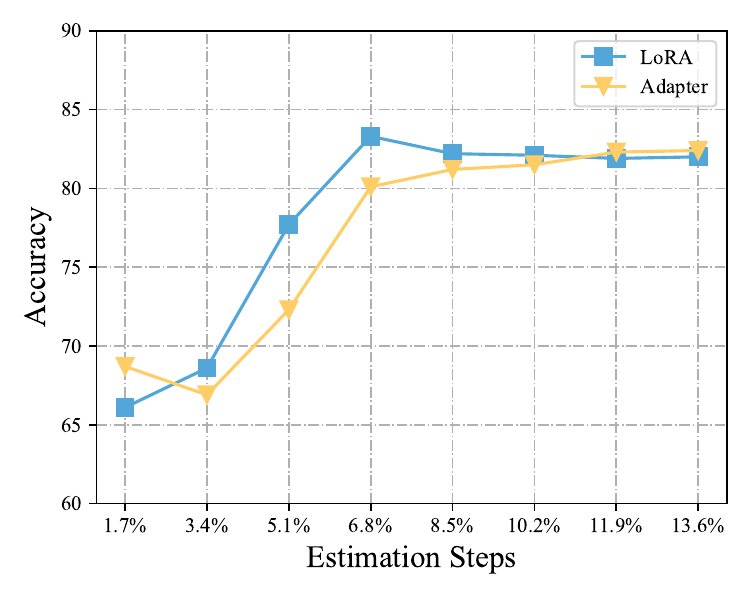}
    \caption{The impact of the estimation steps of early pruning.}
    \label{Appendix-estimation}
\end{figure}

We analyze the impact of the early pruning estimation steps on performance using the BoolQ dataset (Figure \ref{Appendix-estimation}). It is observed that once the estimation steps exceed 6.8\% of the total training steps, further estimation does not lead to performance improvement. This demonstrates that our method can effectively identify redundant parameters in both the foundation model and PEFT modules during the early stage of training.
\subsection{The settings of mask learning penalty}
\label{mask-learning-penalty}
In practice, we keep $\lambda_{A}$ and $\lambda_{F}$ consistent and assess the impact of these hyper-parameters in pilot experiments (Table \ref{lambda-set}). Based on this result, we uniformly set $\lambda_{A}$ and $\lambda_{F}$ to $1\times10^{-4}$ and achieve good task performance in our main experiments.
\begin{table}[H]
\centering
\begin{tabular}{@{}llll@{}}
\toprule
$\lambda_{A}$, $\lambda_{F}$ & SST-2 & QNLI & Avg. \\ \midrule
$1*10^{-2}$ & 95.8 & 91.9 & 93.85          \\
$1*10^{-3}$ & 95.9 & 93.5 & 94.70          \\
$1*10^{-4}$ & 95.9 & 93.6 & \textbf{94.75} \\
$1*10^{-5}$ & 95.6 & 91.9 & 93.75          \\ \bottomrule
\end{tabular}
\caption{The impact of $\lambda_{A}$ and $\lambda_{F}$ on the performance of tasks.}
\label{lambda-set}
\end{table}

\begin{table*}[b]
\centering
\scalebox{0.85}{
\begin{tabular}{@{}cccccccccc@{}}
\toprule
Method & Dataset                              & MNLI & QNLI & QQP & SST-2 & ReCord & WiC & BoolQ & MultiRC \\ \midrule
\multirow{9}{*}{LoRA}    & \multicolumn{1}{c|}{Estimation Steps} & 1000 & 1000 & 1000 & 800 & 2000 & 680 & 400 & 600  \\
       & \multicolumn{1}{c|}{Rank}            & \multicolumn{8}{c}{8}                                      \\
       & \multicolumn{1}{c|}{$\rho_{M}$}         & \multicolumn{8}{c}{75\%}                                   \\
       & \multicolumn{1}{c|}{$\rho_{R}$}         & \multicolumn{8}{c}{50\%}                                   \\
            & \multicolumn{1}{c|}{Estimation lr}    & 3e-4 & 3e-4 & 3e-4 & 3e-4  & 3e-4   & 3e-4 & 3e-4  & 3e-4    \\
     & \multicolumn{1}{c|}{Fine-Tuning lr}   & 3e-4 & 3e-4 & 3e-4 & 3e-4  & 3e-4   & 3e-4 & 3e-4  & 3e-4    \\
       & \multicolumn{1}{c|}{Batch Size}      & 32   & 32   & 32  & 32    & 32     & 16  & 32    & 16      \\
       & \multicolumn{1}{c|}{Sequence Length} & 128  & 128  & 128 & 128   & 256    & 128 & 128   & 384     \\
       & \multicolumn{1}{c|}{\# Epochs}       & 5    & 5    & 5   & 10    & 5      & 50  & 20    & 20      \\ \midrule
\multirow{7}{*}{Adapter} & \multicolumn{1}{c|}{Estimation Steps} & 1000 & 1000 & 1000 & 800 & 2000 & 680 & 400 & 1000 \\
       & \multicolumn{1}{c|}{Rank}            & \multicolumn{8}{c}{8}                                      \\
       & \multicolumn{1}{c|}{$\rho_{M}$}         & \multicolumn{8}{c}{25\%}                                   \\
       & \multicolumn{1}{c|}{$\rho_{R}$}         & \multicolumn{8}{c}{50\%}                                   \\
       & \multicolumn{1}{c|}{Estimation lr}    & 6e-4 & 8e-4 & 3e-4 & 6e-4  & 6e-4   & 3e-4 & 6e-4  & 7e-4    \\
        & \multicolumn{1}{c|}{Fine-Tuning lr}   & 4e-4 & 3e-4 & 3e-4 & 3e-4  & 3e-4   & 1e-4 & 6e-4  & 5e-4    \\
       & \multicolumn{1}{c|}{Batch Size}      & 32   & 32   & 32  & 32    & 32     & 16  & 32    & 16      \\
       & \multicolumn{1}{c|}{Sequence Length} & 128  & 128  & 128 & 128   & 256    & 128 & 128   & 384     \\
       & \multicolumn{1}{c|}{\# Epochs}       & 5    & 5    & 5   & 10    & 5      & 50  & 20    & 20  \\ \bottomrule   
\end{tabular}
}
\caption{Hyperparameters for NLU Tasks.}
\label{glue-setting}
\end{table*}

\begin{table*}[b]
\centering
\scalebox{0.85}{
\begin{tabular}{@{}cc|cccccc@{}}
\toprule
Method                & Dataset          & OpenBookQA & PIQA    & ARC-E   & ARC-C   & SciQ    & WebQs   \\ \midrule
\multirow{9}{*}{LoRA} & Estimation Steps & 1 Epoch    & 1 Epoch & 1 Epoch & 1 Epoch & 1 Epoch & 1 Epoch \\
                      & Rank             & \multicolumn{6}{c}{8}                                        \\
                      & $\rho_{M}$          & \multicolumn{6}{c}{50\%/50\%}                                \\
                      & $\rho_{R}$          & \multicolumn{6}{c}{50\%/25\%}                                \\
                      & Estimation lr    & 3e-4       & 3e-4    & 3e-4    & 3e-4    & 3e-4    & 3e-4    \\
                      & Fine-Tuning lr   & 3e-4       & 3e-4    & 3e-4    & 3e-4    & 3e-4    & 3e-4    \\
                      & Batch Size       & 64         & 64      & 64      & 64      & 64      & 64      \\
                      & Sequence Length  & 128        & 128     & 128     & 128     & 128     & 128     \\
                      & \# Epochs        & 10         & 10      & 10      & 10      & 10      & 10      \\ \bottomrule
\end{tabular}
}
\caption{Hyperparameters for QA Tasks on OPT-1.3B.}
\label{qa-setting-13}
\end{table*}

\begin{table*}[b]
\centering
\scalebox{0.85}{
\begin{tabular}{@{}cc|cccccc@{}}
\toprule
Method                & Dataset          & OpenBookQA & PIQA    & ARC-E   & ARC-C   & SciQ    & WebQs   \\ \midrule
\multirow{9}{*}{LoRA} & Estimation Steps & 1 Epoch    & 1 Epoch & 1 Epoch & 1 Epoch & 1 Epoch & 1 Epoch \\
                      & Rank             & \multicolumn{6}{c}{8}                                        \\
                      & $\rho_{M}$          & \multicolumn{6}{c}{50\%/75\%}                                \\
                      & $\rho_{R}$          & \multicolumn{6}{c}{25\%/50\%}                                \\
                      & Estimation lr    & 3e-4       & 3e-4    & 3e-4    & 3e-4    & 3e-4    & 3e-4    \\
                      & Fine-Tuning lr   & 3e-4       & 3e-4    & 3e-4    & 3e-4    & 3e-4    & 3e-4    \\
                      & Batch Size       & 32         & 32      & 32      & 32      & 32      & 32      \\
                      & Sequence Length  & 128        & 128     & 128     & 128     & 128     & 128     \\
                      & \# Epochs        & 10         & 10      & 10      & 10      & 10      & 10      \\ \bottomrule
\end{tabular}
}
\caption{Hyperparameters for QA Tasks on OPT-6.7B.}
\label{qa-setting-67}
\end{table*}
\end{document}